\documentclass{ieeetj}
\usepackage{cite}
\usepackage{amsmath,amssymb,amsfonts}
\usepackage{algorithmic}
\usepackage{graphicx,color}
\usepackage{textcomp}
\usepackage{xcolor}
\usepackage{hyperref}

\usepackage{subfig}
\usepackage{graphicx}
\graphicspath{ {./figure/} }
\hypersetup{hidelinks}
\usepackage{algorithm,algorithmic}
\def\BibTeX{{\rm B\kern-.05em{\sc i\kern-.025em b}\kern-.08em
    T\kern-.1667em\lower.7ex\hbox{E}\kern-.125emX}}
\AtBeginDocument{\definecolor{tmlcncolor}{cmyk}{0.93,0.59,0.15,0.02}\definecolor{NavyBlue}{RGB}{0,86,125}}

\def\OJlogo{\vspace{-4pt}$<$Society logo(s) and publication title will appear here.$>$}
\def\seclogo{\vspace{10pt}$<$Society logo(s) and publication title will appear here.$>$}

\def\authorrefmark#1{\ensuremath{^{\textbf{#1}}}}

\begin{document}
\receiveddate{XX Month, XXXX}
\reviseddate{XX Month, XXXX}
\accepteddate{XX Month, XXXX}
\publisheddate{XX Month, XXXX}
\currentdate{XX Month, XXXX}
\doiinfo{XXXX.2022.1234567}

\markboth{}{Author {et al.}}

\title{S5-HES Agent: Society 5.0-driven Agentic Framework to Democratize Smart Home Environment Simulation}


\author{Akila Siriweera\authorrefmark{1}, Member, IEEE,
        Janani Rangila\authorrefmark{2}, Student Member, IEEE,
        Keitaro Naruse\authorrefmark{1}, Member, IEEE,
        Incheon Paik\authorrefmark{1}, Senior Member, IEEE,
        and Isuru Jayanada\authorrefmark{2}, Student Member, IEEE}

\affil{The University of Aizu, Aizu Wakamatsu, Fukushima, Japan}
\affil{The KD University, Rathmalana, Colombo, Sri Lanka}

\begin{abstract}
The smart home is a key 
domain within the Society 5.0 vision for a human-centered society.
Smart home technologies rapidly evolve, and research should diversify while remaining aligned with Society 5.0
objectives.
Democratizing smart home research would engage a broader community of innovators beyond traditional
limited experts.
This shift necessitates inclusive simulation frameworks that support research across diverse fields in industry and academia.
However, existing smart home simulators require significant technical expertise, offer limited adaptability, and lack automated
evolution, thereby failing to meet the holistic needs of Society 5.0.
These constraints impede researchers from efficiently
conducting simulations and experiments
for
security,
energy, 
health,
climate,
and socio-economic research.
To address these challenges, this paper presents the Society 5.0-driven Smart Home Environment Simulator Agent (S5-HES Agent), an agentic simulation framework that transforms traditional smart home simulation through autonomous AI orchestration.
The framework coordinates specialized agents through interchangeable large language models (LLMs), enabling natural-language-driven end-to-end smart home simulation configuration without programming expertise.
A retrieval-augmented generation (RAG) pipeline with semantic, keyword, and hybrid search retrieves
smart home
knowledge.
Comprehensive evaluation on S5-HES Agent demonstrates that the RAG pipeline achieves near-optimal retrieval fidelity,
simulated device behaviour and threat scenarios align with real-world IoT datasets, and
simulation engine scales predictably across home configurations, establishing a stable foundation for Society 5.0 smart home research. 
Source code is available under the MIT License at https://github.com/AsiriweLab/S5-HES-Agent.
\end{abstract}



\maketitle

\section{INTRODUCTION}

\IEEEPARstart{T}{he} Japanese Cabinet Office's vision for Society 5.0 describes a human-centered society in which cyber and physical systems are integrated to improve social well-being alongside economic progress \cite{CabinetOffice, 10752374, siriweera2025autobda}. 
The smart home is the primary residence where people interact with interconnected devices to maintain and improve their quality of life.
As smart home ecosystems expand in scope and complexity, spanning various protocols (such as MQTT and CoAP), device categories, and evolving threats, the research community requires tools that can simulate these environments at scale while remaining reproducible across disciplines 
\cite{11229932}.

Smart home research has often progressed within disciplinary boundaries.
Research domains, such as security emphasizes threats \cite{mbc1-1h68-22,garcia2020iot }, energy focuses on optimization \cite{galasso2023smart, amer2024development, gaikwad2025smart}, and health studies well-being \cite{sadhwani2023fleet2d}. 
However, the Society 5.0 aims holistic and integrated solution
\cite{CabinetOffice, 10752374, siriweera2025autobda}.
Moreover, democratization of smart home ecosystems opens the domain
beyond limited experts \cite{siriweera2025autobda}. Home environment simulation (HES) solutions that are inclusive by design and avoid prohibitive technical constraints are essential to achieve the Society 5.0 vision \cite{CabinetOffice, 10752374}.

Real-world HES testbeds are expensive to deploy, difficult to reproduce across laboratories, and constrained by privacy regulations that limit data sharing \cite{spoladore2017semantic, alshammari2017openshs, madolia2023right}.
HES platforms mitigate these constraints by enabling controlled, repeatable experiments that can configure device topologies, inject threats, vary environmental parameters, and generate labeled datasets at scale. 
Because such HES data frequently support downstream tasks (such as digital modeling and multidisciplinary research), the fidelity and configurability of the HES are critical research requirements.

Despite their utility, widely used HES and datasets remain difficult to configure and extend.
Incorporating new categories or threat types typically requires manual implementation by domain experts, which narrows the user base and conflicts with Society 5.0's inclusivity goals.
Limitations appear in the datasets used for evaluation and model development \cite{s150614162, 10775126,
bolzani2006domotics, van2009iss, ghayvat2015simulation,
chen2016smartsim, tanantong2017towards, amer2024development,
kumar2023framework, sadhwani2023fleet2d, gaikwad2025smart,
spoladore2017semantic, alshammari2017openshs, madolia2023right,
galasso2023smart, 10599909, sarvaiya2025simulating,
s22218109,
qrq9-8469-24,
y9de-qj71-25,
garcia2020iot,
mbc1-1h68-22,
r7v2-x988-19}
are static captures produced from specific testbeds under fixed conditions. None of them is configurable, and manual labeling is both labor-intensive and error-prone. No single dataset spans the breadth of scenarios needed for Society 5.0.
These constraints motivate a configurable, reproducible HES framework that produces realistic and labeled data tailored to diverse research questions.

Recent advances in large language models (LLMs) and agentic AI architectures provide pragmatic solutions to address these constraints \cite{brown2020language}.
LLMs can map natural-language intent to structured configurations, and multi-agent systems can execute complex workflows. 
Retrieval-augmented generation (RAG) leverages retrieved domain evidence to further ground LLM outputs \cite{lewis2020retrieval}, improving factual consistency.
However, these techniques have not been effectively integrated into smart home simulation frameworks.

Therefore, the existing constrained smart home domain leaves a threefold gap. 
%
First, existing simulators do not provide a natural-language-driven end-to-end configuration for non-expert users to build complex environments without programming.
Second, they do not provide ground simulation parameters in retrieved domain knowledge, such as specs, threats, and literature, to support behavioral realism.
Third, none provide automatic ground-truth labeling with configurable taxonomy depth, from binary (benign/malicious) division to fine-grained and labeled MITRE ATT\&CK \cite{strom2018mitre}.

To address these gaps, we present the Society 5.0-driven Smart Home Environment Simulator Agent (S5-HES Agent), an agentic simulation framework that supports automated orchestration of end-to-end smart home simulation. The framework adopts a three-layer architecture:
a \textit{presentation layer} provides web and API interfaces,
a \textit{cognitive layer} orchestrates multi-agents via RAG pipeline,
and an \textit{engine layer} executes simulation and integrity.
%
%
To the best of our knowledge, S5-HES Agent is the first automated, AI-augmented framework with agentic orchestration, designed to democratize smart HES domain.
%
Our key contributions;
\begin{itemize}
    \item An agentic RAG framework for smart home simulation that coordinates specialized agents through interchangeable LLMs and retrieves domain knowledge via hybrid search 
    academic, specification, and threat documents.
    
    \item A configurable simulation engine that generates
    IoTs
    and MITRE ATT$\&$CK-mapped threats with automatic ground-truth labeling at multiple taxonomy levels.
    
    \item A dual-mode interaction paradigm for LLM
    and No-LLM assisted configuration, lowering the expertise barrier while preserving expert control.
    
    \item An end-user-friendly end-to-end simulation framework that enables users to configure and run simulations with zero programming expertise.
    
    \item Alignment with Society 5.0;
    enabling an interdisciplinary and inclusive smart home research ecosystem.
\end{itemize}

This paper is organized as follows. Section \ref{sec_2} reviews related work. Section \ref{sec_3} outlines preliminaries. Section \ref{sec_4} presents the modeled S-HES Agent. 
In Section \ref{sec_6}, we present the evaluation results. Section \ref{sec_7} concludes the paper.


\section{Related Works}
\label{sec_2}
\noindent This section investigates the smart HES frameworks that facilitate various levels of automation: autonomous (supports end-to-end automated workflow with minimal manual intervention), automated (requires some configuration but runs automated processes), and manual (requires extensive programming or manual configuration). Selected literature works belonging to recent major publications and well-known HES frameworks 
\cite{s150614162, 10775126,
bolzani2006domotics, van2009iss, ghayvat2015simulation,
chen2016smartsim, tanantong2017towards, amer2024development,
kumar2023framework, sadhwani2023fleet2d, gaikwad2025smart,
spoladore2017semantic, alshammari2017openshs, madolia2023right,
galasso2023smart, 10599909, sarvaiya2025simulating}. 
We conducted a subjective survey based on our research objective.
We observed the following objective taxonomy after scrutiny, which helps maintain the objectivity of the subjective investigation.

We observed differences in holistic capabilities across domains and technical features. And noted three inclusive classes based on the awareness of domain scope, experimental capabilities, and knowledge integration. Sub-classes of inclusive are: based on the domain scope 'single-residence' or 'multi-residence,' based on the experimental capabilities 'custom,' 'scalability,' or 'reproducibility,' and according to the knowledge integration, solutions with 'none,' 'static,' or 'adaptive' knowledge-driven features. Finally, agile classes are 'manual,' 'automated,' and 'autonomous.' 

Based on that classification, we observed the 
six main groups (G1 through G6) are identified based on their distinctive capability patterns across domain, experimentation, knowledge-driven features, and workflow automation.

The first group (G1), comprising works {\cite{s150614162}} and {\cite{10775126}}, represents early-stage frameworks with basic functionality. These solutions are limited to single-residence domains, offer minimal experimental capabilities, lack knowledge-driven features, and operate entirely through manual workflows.

Building upon this foundation, G2 frameworks {\cite{bolzani2006domotics, van2009iss, ghayvat2015simulation}} maintain a single-residence focus while introducing reproducibility features. Although these solutions still lack knowledge-driven capabilities and rely primarily on manual workflows, they represent incremental improvements in experimental rigor.
Frameworks in G3 {\cite{chen2016smartsim, tanantong2017towards, amer2024development}} demonstrate further advancements by incorporating both scalability and reproducibility for single-residence domains. While knowledge-driven features remain absent, these solutions show notable progression in workflow automation through mixed manual and automated approaches.

The fourth group (G4), represented by works {\cite{kumar2023framework, sadhwani2023fleet2d}} and {\cite{gaikwad2025smart}}, achieves fully automated workflows for single-residence scenarios with scalability features. Despite lacking knowledge-driven capabilities, these frameworks exhibit more sophisticated technical implementations than earlier groups.
In contrast, G5 {\cite{spoladore2017semantic, alshammari2017openshs, madolia2023right}}  prioritizes flexibility in experimental design by introducing custom experiment abilities for single-residence domains. However, they maintain manual workflows and lack knowledge-driven features.

Finally, G6 frameworks {\cite{galasso2023smart, 10599909, sarvaiya2025simulating}} represent the most advanced pre-existing solutions, incorporating custom experimentation capabilities and static knowledge-driven features through AI and language model technologies. Although these frameworks operate through automated workflows within single-residence domains, they still require semantic knowledge in smart home systems for effective utilization.

S5-HES Agent possesses unique characteristics throughout the taxonomy: multi-residence domain support, comprehensive experimental capabilities (custom, scalability, reproducibility), adaptive knowledge-driven features, and support for manual, automated, and semi-autonomous workflows. From a democratization perspective, its architecture is relatively more flexible for adapting/scaling to diverse research needs across multiple research environments.


In summary, the groups demonstrate evolutionary progression in smart home simulation capabilities. 
Nevertheless, from a democratization standpoint, the accessibility remains constrained by domain limitations and technical requirements. 
G6 frameworks are more holistic and inclusive within the single-residence context; however, our proposed solution extends these capabilities to multi-residence environments with adaptive knowledge systems. 
Furthermore, while G4 and G6 are more agile with automated workflows, they require semantic knowledge in smart home systems. 
Among all frameworks, S5-HES Agent is the only multi-residence approach that supports autonomous, thorough end-to-end automated workflows without requiring semantic knowledge in simulation configuration or automation. This advantage significantly improves democratized access to the smart home research domain.

\section{Preliminary}
\label{sec_3}
This section introduces the motivation scenario and problem formulation for the proposed framework. 
In Section \ref{sec3_sub1}, we discuss the S5-HES research scenario. Section \ref{sec3_sub2} presents the problem definition and six key challenges identified from the motivating scenario.

\subsection{Smart Home Research Scenario}
\label{sec3_sub1}
A research group called HES-Research is conducting studies for Society 5.0-driven smart residential environments. The research team requires the generation of datasets for machine learning model development across diverse smart home configurations. The research infrastructure operates on a simulation-based model in which all environmental generation is simulated, eliminating the need for physical testbeds and hardware infrastructure. Each simulated resident can be configured with multiple functional zones (such as living room, bedroom, kitchen, bathroom, office, garage, and garden) drawn from many room types, and with device clusters comprising sensors, actuators, controllers, and communication hubs spanning various device categories.

HES-Research must address three critical concerns. First, ensuring flexibility through configurable home environments supporting diverse residence types from studio apartments to multi-story mansions with varying device populations. Second, guaranteeing reproducibility by enabling researchers to regenerate identical simulation scenarios across different institutions and time periods, independent of hardware. Third, supporting both manual workflows (expert-driven configuration of specific scenarios) and AI-assisted workflows. Therefore, HES-Research requires a simulation framework that supports flexibility (customization), reproducibility (deterministic generation with seed control and experiment versioning), and accessibility (reduced programming requirements through natural-language interfaces).

\subsection{Problem Definition}
\label{sec3_sub2}
Based on the scenario described in Section \ref{sec3_sub1}, we formalize the smart home environment simulation problem as a unified definition comprising six requirements that existing approaches fail to address jointly.

\noindent \textbf{Definition 1:} \textit{Smart Home Environment Simulation Problem:}
Let $\mathcal{H}$, $\mathcal{R}$, $\mathcal{D}$, and $\mathcal{P}$ denote the sets of home types, room configurations, device types, and inhabitant models, respectively. A smart home environment simulation framework must jointly satisfy six requirements: 

(1) \textit{Environment Flexibility} supporting configurable generation $\mathcal{F}_{home} : (\mathcal{H}, \mathcal{R}, \mathcal{D}, \theta) \rightarrow \mathcal{C}$ across heterogeneous residence types and device populations; 

(2) \textit{Multi-Inhabitant Presence Modeling} correlating device activation with inhabitant presence through $\mathcal{A}_d(t) = \max_{i} \mathbb{I}_{present}(p_i, t) \cdot f_{activity}(p_i, t)$, where activity likelihoods are governed by time-dependent Markov transition matrices; 

(3) \textit{Threat Scenario Customization} enabling configurable attack injection $\mathcal{T}_{inject} : (t_i, D_{target}, \tau_{window}, \gamma) \rightarrow \mathcal{S}_{attack}$ with MITRE ATT\&CK-mapped threat types, target device selection, timing, and intensity control; 

(4) \textit{Knowledge Extensibility} providing runtime-expandable knowledge retrieval $K_{RAG} : (Q, KB) \rightarrow \{(doc_i, S_{RRF_i})\}_{i=1}^{k}$ via hybrid search combining semantic similarity and keyword matching through reciprocal rank fusion; 

(5) \textit{Reproducibility} guaranteeing deterministic output equivalence $Sim(\xi, \Omega)\{t_1\} \equiv Sim(\xi, \Omega)\{t_2\}, \forall t_1, t_2$ through seeded generation with complete configuration export; and 

(6) \textit{Accessibility} lowering expertise barriers through natural language interfaces with multi-agent orchestration $NL_{pipeline} : Q_{natural} \xrightarrow{\text{Agents}} C_{sim}$, where specialized agents translate natural language input into HES configs.

\section{Proposed Method}
\label{sec_4}
This section presents the S5-HES Agent designed to address the six requirements identified in Definition 1. 
Section \ref{sec4_sub2} presents the system architecture with implementation details. Sections \ref{sec4_sub3}  \ref{sec4_sub8} detail the solutions to each requirement.




\subsection{System Architecture}
\label{sec4_sub2}

\begin{figure}[t]
\centering
\includegraphics[width=\columnwidth]{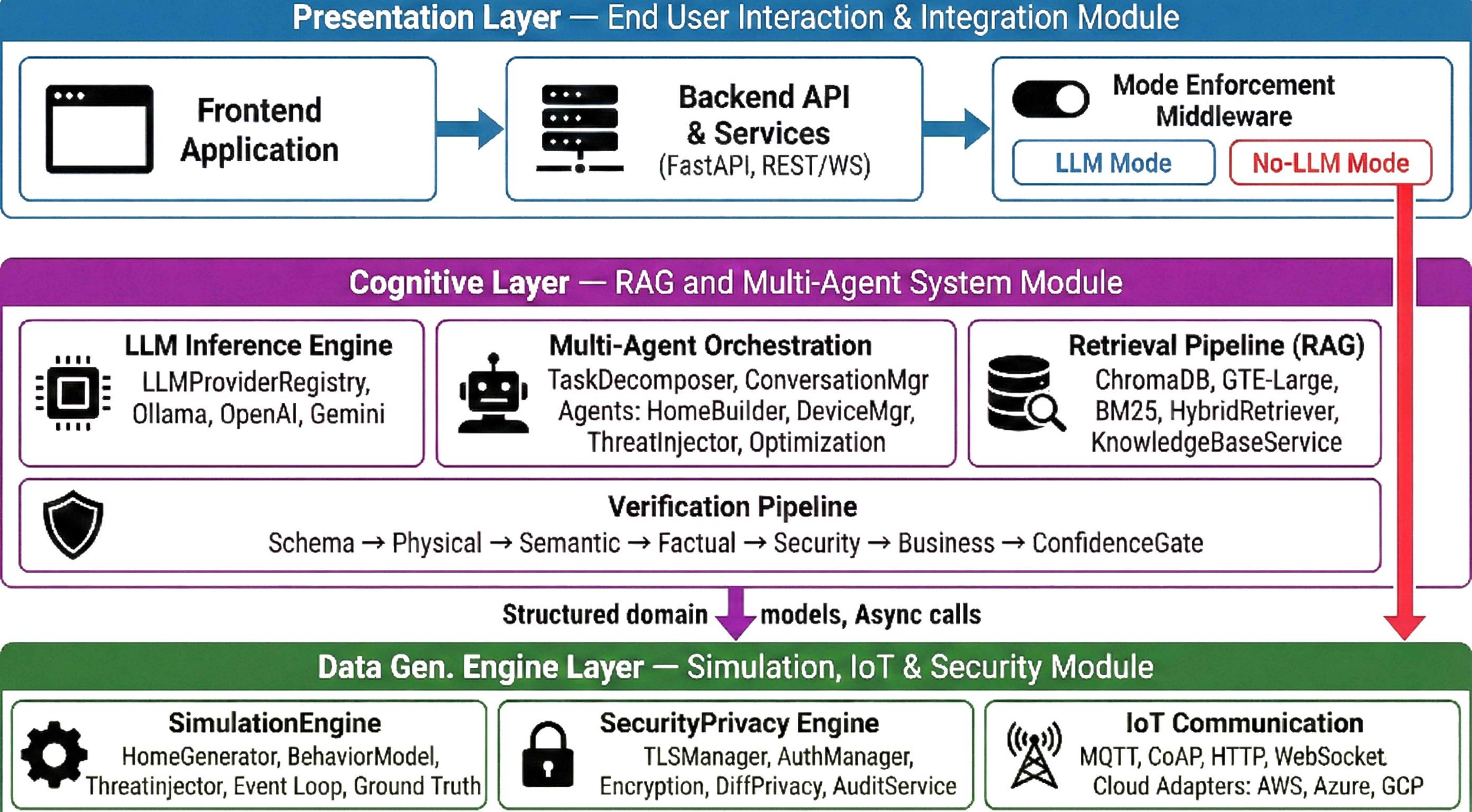}
\caption{S5-HES: Normative System architecture}
\label{fig:sys_arch}
\end{figure}

Fig. \ref{fig:sys_arch} presents the system architecture instantiating the reference architecture into deployable components. The \textit{Presentation Layer} comprises a Vue.js 3 frontend connecting via REST, SSE, and WebSocket protocols to a FastAPI backend with Pydantic v2 request validation. The ModeEnforcementMiddleware enables dual-mode operation: LLM mode provides full AI-assisted access, while No-LLM mode restricts endpoints to simulation, RAG search, and data retrieval. Three input modalities serve different expertise levels: chat input for conversational configuration, visual builders for manual setup, and API calls for programmatic access.

The \textit{Cognitive Layer} contains four components: (1) an LLMEngine with a LLMProviderRegistry supporting runtime switching among Ollama (local), OpenAI, and Gemini; (2) an Orchestrator coordinating four specialized agents (HomeBuilder, DeviceManager, ThreatInjector, Optimization) via a TaskDecomposer with ConversationMgr maintaining session context; (3) a Retrieval Pipeline combining ChromaDB with GTE-Large embeddings (1024 dimensions) and BM25 indexing through a HybridRetriever over 20,306 pre-indexed document chunks with provenance tracking; and (4) a Verification Pipeline chaining schema, physical, semantic, factual, security, and business rule validators through a ConfidenceGate (thresholds: 0.85 automatic approval, 0.70 human review).

The \textit{Data Generation Engine Layer} houses the SimulationEngine supporting 118 device types with Markov-chain activity modeling, 22 threat types with six-phase attack lifecycles, and time-compressed event generation at 1440$\times$. The SecurityPrivacyEngine provides TLS transport security, JWT/OAuth authentication, AES-256-GCM encryption, and differential privacy. IoT Communication implements protocol handlers for MQTT, CoAP, HTTP, and WebSocket with cloud platform adapters for AWS IoT, Azure IoT, and GCP.

\subsection{Configurable Home Builder: Solution to Definition 1.1}
\label{sec4_sub3}

The Home Generator implements $\mathcal{F}_{home}$ through a template-based system with five residence types $\mathcal{T} = \{\tau_1, \ldots, \tau_5\}$, each specified as $\tau_i = (|\mathcal{R}_i|, S_i, \delta_i)$ defining room count, floor area (m$^2$), and device density bounds. 

Generation proceeds through three stages. Template selection identifies the matching residence type. Room instantiation generates rooms by sampling from probability distributions over 15 room types (living room, bedroom, kitchen, bathroom, office, garage, hallway, etc.):
\begin{equation}
\mathcal{R}_{home} = \{r_j : j \in [1, |\mathcal{R}(\tau)|],\ \text{type}(r_j) \sim P_{room}(\tau)\}
\label{eq:room}
\end{equation}
where each room receives dimensions, coordinates, and adjacency relationships. Device placement distributes 118 device types across 16 categories (security, lighting, climate, entertainment, kitchen, health, energy, etc.) using compatibility-weighted density functions:
\begin{equation}
\mathbb{E}[|D_c(r_j)|] = \lambda_c \cdot \text{area}(r_j) \cdot \text{compat}(c, \text{type}(r_j))
\label{eq:device_density}
\end{equation}
where $\lambda_c$ is the base density for category $c$ and $\text{compat}(\cdot) \in [0,1]$ encodes category-room compatibility ensuring realistic distributions (e.g., security cameras at entrances, kitchen appliances in cooking spaces). Each device specifies $\text{spec}(d) = (\text{protocols}, \text{power}, \text{states}, \text{transitions})$ where $\text{protocols} \subseteq \mathcal{P}$ covers 12 supported communication protocols (WiFi, Zigbee, Z-Wave, Bluetooth, BLE, Matter, Thread, MQTT, HTTP, CoAP, Ethernet, Modbus), enabling protocol-accurate traffic generation. The output $\mathcal{C} = (\mathcal{R}_{home}, D_{placed}, G_{network}, M_{behavior})$ produces deterministic configurations for identical seeds, simultaneously addressing reproducibility.


\subsection{Multi-Inhabitant Behavior: Solution to Definition 1.2}
\label{sec4_sub4}

The behavior engine models inhabitants $\mathcal{P} = \{p_1, \ldots, p_r\}$ with attributes including type $\in \{\text{adult, child, elderly, teenager, pet}\}$, schedule parameters $\sigma_i$ (wake/sleep times, work hours, work-from-home flag), and tech-savviness $\pi_i \in [0,1]$. Presence probability derives from schedule-based distributions:
\begin{equation}
\Pr(\mathbb{I}_{present}(p_i, t) = 1) = P_{schedule}(\sigma_i, t) \cdot (1 + \epsilon_i(t))
\label{eq:presence}
\end{equation}
where $\epsilon_i(t) \sim \mathcal{N}(0, \sigma^2_{var})$ introduces individual variation. Activity sequences follow a time-inhomogeneous Markov Chain over 16 activity states (sleeping, waking up, working, cooking, entertainment, etc.) with transition matrices varying across four diurnal periods:
\begin{equation}
\Pr(s_{t+1} = s' \mid s_t = s, p_i) = M^{(\tau(t))}_{s,s'}(p_i)
\label{eq:markov}
\end{equation}
where $\tau(t) \in \left\{
\begin{array}{l}
\text{morning}\ (06\text{  }09), \text{daytime}\ (09\text{  }17),\\
\text{evening}\ (17\text{  }21), \text{night}\ (21\text{  }06)
\end{array}
\right\}$, capturing realistic daily rhythms (e.g., high sleeping-to-personal-care transitions in morning, elevated entertainment in evening). Device interaction likelihood combines temporal, proficiency, and contextual factors:
\begin{equation}
f_{activity}(p_i, t) = \phi_{temporal}(t) \cdot \phi_{tech}(\pi_i) \cdot \phi_{context}(s_t^{(i)})
\label{eq:activity}
\end{equation}
where $\phi_{temporal}(t)$ captures time-of-day patterns, $\phi_{tech}(\pi_i)$ scales interaction frequency by tech-savviness, and $\phi_{context}(s_t^{(i)})$ links current activity to relevant devices. The complete device activation aggregates individual contributions:
\begin{equation}
\mathcal{A}_d(t) = \max_{i=1}^{r}\, \chi_i(d) \cdot \mathbb{I}_{present}(p_i, t) \cdot f_{activity}(p_i, t)
\label{eq:activation}
\end{equation}
where $\chi_i(d) \in \{0,1\}$ indicates device-inhabitant compatibility based on device category and room type. An occupancy model enforces room-level capacity limits (e.g., bathroom: 1, kitchen: 4, living room: 8) preventing unrealistic overcrowding.

\subsection{Threat Injection Engine: Solution to Definition 1.3}
\label{sec4_sub5}

The Threat Injector implements 22 attack types organized across seven categories mapped to MITRE ATT\&CK technique identifiers. Each attack specification $\text{spec}(t_i) = (\text{pattern}, \text{protocol}, \text{signature}, \text{label}_{ATT\&CK})$ captures generation parameters and progresses through a six-phase lifecycle: reconnaissance, initial access, execution, persistence, exfiltration, and cleanup, generating temporally realistic attack sequences. Target devices are selected by vulnerability profiles $D_{target} = \{d \in \mathcal{D} : \text{vuln}(d, t_i) > \theta_{vuln}\}$, enabling targeting of specific device categories or broad network reconnaissance.

The intensity parameter $\gamma \in [0,1]$ modulates attack characteristics:
\begin{equation}
\begin{aligned}
\text{rate}(t_i, \gamma) = \text{rate}_{base}(t_i) \cdot (1 + \gamma \cdot k_{rate}), \quad \text{stealth}(t_i, \gamma) = \\ \text{stealth}_{base}(t_i) \cdot (1 - \gamma \cdot k_{stealth})
\label{eq:intensity}
\end{aligned}
\end{equation}
enabling systematic generation from subtle ($\gamma < 0.3$) to obvious ($\gamma > 0.7$) signatures for detection algorithm evaluation. Protocol-specific packet crafting ensures realistic network behavior with intensity-scaled parameters (e.g., brute force attempt counts). A ground truth Labeler automatically annotates all traffic:
\begin{equation}
\text{label}(\text{pkt}) = \begin{cases} (t_i,\, \text{label}_{ATT\&CK}(t_i),\, \text{conf}) & \text{if } \text{pkt} \in \mathcal{S}_{attack} \\ (\text{benign},\, \emptyset,\, 1.0) & \text{otherwise} \end{cases}
\label{eq:label}
\end{equation}
producing labeled datasets for supervised machine learning without manual annotation.


\subsection{RAG pipeline: Solution to Definition 1.4}
\label{sec4_sub6}

The Retrieval Pipeline maintains a knowledge base $KB = \{(\text{chunk}_i, \text{emb}_i, \text{meta}_i)\}$ of 20,306 document chunks embedded using GTE-Large ($\mathbb{R}^{1024}$) with chunk size of 512 tokens and 50-token overlap. Three specialized adapters handle domain-specific preprocessing: Academic (research papers, standards documents) for methodology guidance, Threat (CVE reports, ATT\&CK descriptions) for attack pattern specifications, and Device (datasheets, protocol specifications) for device behavior modeling. Hybrid search combines semantic cosine similarity with BM25 keyword scoring through Reciprocal Rank Fusion:
\begin{equation}
S_{RRF}(Q, d) = \frac{w_{sem}}{\kappa + \text{rank}_{sem}(d)} + \frac{w_{kw}}{\kappa + \text{rank}_{kw}(d)}
\label{eq:rrf}
\end{equation}
where $w_{sem} = 0.7$, $w_{kw} = 0.3$, and $\kappa = 60$, addressing the limitation of pure semantic search on technical terminology (e.g., CVE identifiers, protocol names). The retrieval function returns top-$k$ results:
\begin{equation}
\begin{aligned}
K_{RAG}(Q, KB) = \text{Top-}K\big(\{(\text{chunk}_i, S_{RRF}(Q, \text{chunk}_i)) : \\ \text{chunk}_i \in KB\},\, k\big)
\label{eq:retrieval}
\end{aligned}
\end{equation}
Runtime document ingestion via REST endpoints accepts PDF, Markdown, and JSON formats, enabling researchers to incorporate new CVE reports and device specifications without code modification.

\subsection{Deterministic Simulation: Solution to Definition 1.5}
\label{sec4_sub7}

Reproducibility is ensured through seed-controlled simulation where all stochastic components derive from seeded generators $RNG_i = PRNG(\xi + \text{salt}_i)$, with separate instances for device placement, traffic timing, behavior transitions, and attack sequencing preventing cross-component interference. The parameter set captures complete configuration:
\begin{equation}
\Omega = (\theta_{home},\, \theta_{residents},\, \theta_{threats},\, \theta_{duration},\, \theta_{protocols})
\label{eq:params}
\end{equation}
where $\theta_{duration}$ sets simulation timespan with default time compression of 1440$\times$ (24 simulated hours per real minute). Three-tier state preservation records configuration state $(\xi, \Omega, \text{version}, \text{timestamp})$, generation outputs (home configuration, behavior schedules, attack timelines, network topology), and runtime artifacts (traffic sequences, event logs, checkpoints). Experiment versioning assigns identifiers $\text{exp\_id} = \text{hash}(\xi, \Omega, \text{version})$ for cross-institutional reproduction. A parameter sweep component supports Cartesian product exploration with deterministic seed derivation $\xi_j = \xi_{base} + j$ and integrity-verified JSON export/import, guaranteeing $Sim(\xi, \Omega)_{t_1} \equiv Sim(\xi, \Omega)_{t_2}$.

\subsection{Multi-Agent Orchestration: Solution to Definition 1.6}
\label{sec4_sub8}

The Cognitive Layer coordinates four specialized agents: HomeBuilderAgent (natural language descriptions $\rightarrow$ home configuration JSON), DeviceManagerAgent (requirement specifications $\rightarrow$ device manifests), ThreatInjectorAgent (threat descriptions $\rightarrow$ attack timelines), and OptimizationAgent (research objectives $\rightarrow$ optimized parameters). Each implements $A_i : (\text{input}_i, \text{context}_i) \rightarrow (\text{output}_i, \text{confidence}_i)$ where context includes RAG-retrieved knowledge and conversation history. The LLMEngine classifies natural language intent via the LLMProviderRegistry supporting runtime provider switching with a no-fallback integrity policy (explicit errors rather than degraded output). For multi-intent queries (e.g., ``Create a family home with cameras and simulate a brute force attack''), the Orchestrator decomposes requests into dependency-ordered task graphs $G_{tasks} = (V_{tasks}, E_{dependencies})$.

A six-stage Verification Pipeline ensures output reliability by aggregating schema, physical, semantic, factual (RAG-based fact checking), security, and business rule validation scores:
\begin{equation}
\text{conf}(\text{output}) = \prod_{i=1}^{6} V_i(\text{output})^{w_i}
\label{eq:confidence}
\end{equation}
The confidence gate routes outputs with score $\geq 0.85$ for automatic execution, $[0.70, 0.85)$ for human review, and $< 0.70$ for rejection with explanatory feedback. Dual-mode operation ensures accessibility: LLM mode enables conversational AI-assisted workflows through the full Cognitive Layer, while No-LLM mode bypasses it entirely, routing requests directly to the Engine Layer via visual builders for deterministic expert control.

\section{Evaluation}
\label{sec_6}
This section presents an evaluation of the S5-HES.
Section \ref{eval_sec_1} describes the evaluation metrics. Section \ref{eval_sec_2} presents the experiment setup. And, evaluation results are presented in Section \ref{eval_sec_3} and Section \ref{eval_sec_4}.
Section \ref{eval_sec_5} conducts discussion.

\subsection{Evaluation Metrics}
\label{eval_sec_1}

The evaluation involves two key dimensions as follows.

\begin{itemize}
    \item RAG pipeline:
        \begin{itemize}   
            \item Retrieval quality is measured using standard information retrieval metrics: Precision at k (P@k), Recall at k (R@k), and normalised Discounted Cumulative Gain (nDCG). Mean reciprocal rank (MRR) and mean average precision (MAP) are compared against established base models, BM25, E5, GTE, and BGE \cite{BM25_10.1561/1500000019, E5_wang2022text, GTE_li2023towards
            }. 
             \item Response quality assessed through Faithfulness (context grounding), Fluency (structural coherence), ROUGE-L (lexical overlap with expected answers) \cite{rouge_lin2004rouge}, and BERTScore (semantic similarity) \cite{bert_zhang2019bertscore} and compares with LLM providers, Llama 3.2-3B,
             Gemini 2.0-Flash,
             and GPT-4o.
        \end{itemize}
    
    \item Data generation engine:
        \begin{itemize}   
            \item Threat scenario fidelity evaluated using two domain-independent metrics: attack behaviour coverage (ABC) validates  MITRE ATT$\&$CK indicator  \cite{strom2018mitre}, and attack lifecycle fidelity (ALF) assesses cyber kill chain phase coverage and ordering \cite{hutchins2011intelligence}. Edge-IIoTset \cite{mbc1-1h68-22}, IoT-23 \cite{garcia2020iot}, and Bot-IoT \cite{r7v2-x988-19} are employed as baselines for labeling attacks.
            \item Device behaviour realism is quantified through message-level similarity (M.SIM) against two datasets, SDHAR-HOME \cite{s22218109} and Logging \cite{qrq9-8469-24}. 
            \item Dataset quality is assessed across seven dimensions (scale, feature count, balance, attack diversity, temporal uniformity, diversity, and taxonomy depth) against three baselines, N-BaIoT \cite{y9de-qj71-25}, IoT-23 \cite{garcia2020iot}, and 
            TON-IoT 
            \cite{9189760}.
            \item Dataset capabilities were studied under three metrics: scalability and diversity. 
        \end{itemize}

\end{itemize}

\subsection{Experimental Setup}
\label{eval_sec_2}
All experiments were conducted on a single workstation equipped with an AMD Ryzen AI 9 HX 370 processor with Radeon 890M integrated graphics (24 CPUs, 2.0 GHz base clock), running Windows 11 with Python 3.13.5. The S5-HES knowledge base comprises 20,306 documents stored in ChromaDB using GTE-Large embeddings. Retrieval experiments use 100 queries spanning smart home knowledge categories. Data generation experiments use a fixed random seed (SEED = 42) for reproducibility, with each experiment session timestamped and archived. All evaluation notebooks and session artefacts are stored under a versioned directory structure to enable full reproducibility.

\subsection{RAG pipeline}
\label{eval_sec_3}
We evaluate the RAG pipeline in two directions. 
First, we assess document retrieval accuracy by comparing the S5-HES retrieval engine against four established embedding and lexical baselines using standard information retrieval metrics. 
Second, we evaluate the downstream quality of responses generated by the RAG pipeline with multiple LLMs.

\subsubsection{Retrieval quality}
The retrieval pipeline is evaluated using 100 queries spanning five smart home knowledge categories (x20 queries), drawn from the S5-HES ground truth corpus. Three S5-HES retrieval variants are compared: Hybrid (combining keyword and semantic search with reciprocal rank fusion), Semantic (GTE-Large embedding similarity), and Keyword (BM25-based lexical matching). 
These are benchmarked against four external baselines (BM25, GTE, E5, and BGE). Performance is measured using set-based metrics (Precision@K, Recall@K), rank-weighted quality (nDCG@K), and position-based summary metrics (MRR, MAP), with K values of 1,5,10,20.
Metrics are computed per query and aggregated as mean ± standard deviation over the 100-queries. 

\begin{figure}[!t]
\vspace{-5mm}
\centering
\subfloat[Precision@K]{
\begin{minipage}[t]{0.31\linewidth} 
\centering 
\includegraphics[scale=0.35]{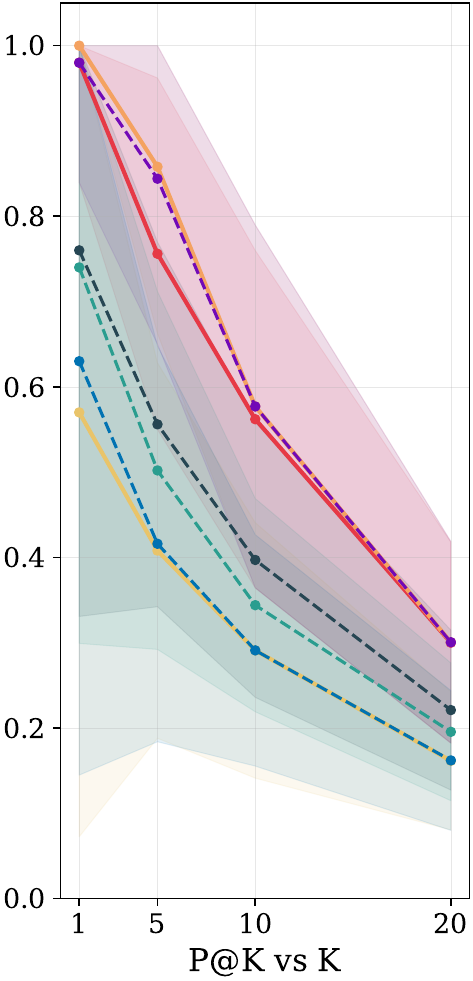} 
\label{eval_fig_1_A}
\end{minipage}%
}%
\subfloat[Recall@K]{
\begin{minipage}[t]{0.30\linewidth} 
\centering 
\includegraphics[scale=0.35]{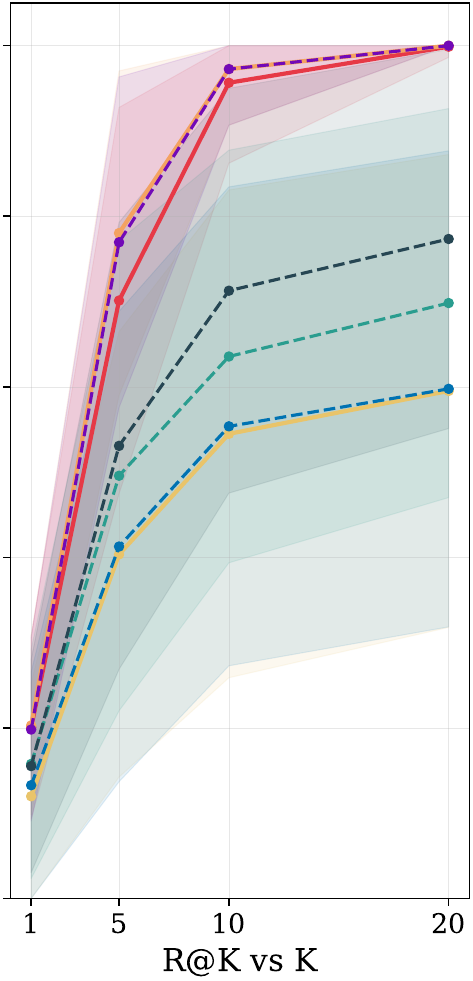}
\label{eval_fig_1_B}
\end{minipage} 
}
\subfloat[nDCG@K]{
\begin{minipage}[t]{0.30\linewidth} 
\hspace{-3.5mm}  
\centering 
\includegraphics[scale=0.35]{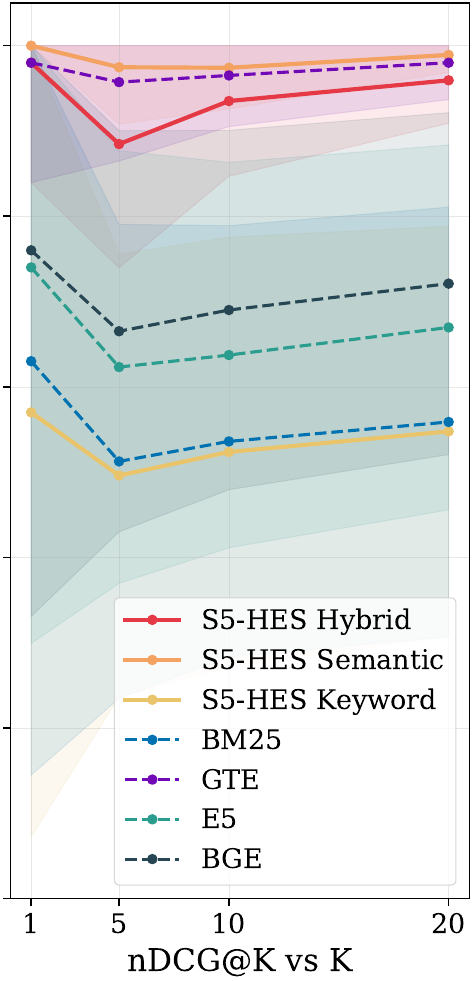}
\label{eval_fig_1_C}
\end{minipage} 
}
\caption{Set-based and rank-weighted performance
\newline (Solid lines for S5-HES variants and dashed lines for baselines. Shaded regions indicate ±1 std over 100 queries.)
}
\vspace{-9mm}
\label{eval_fig_1}
\end{figure}

\begin{figure}[t]
\vspace{-5mm}
\hspace{-5mm}  
\centering
\subfloat[Mean reciprocal rank vs Models]{
\begin{minipage}[t]{0.50\linewidth} 
\centering 
\includegraphics[scale=0.5]{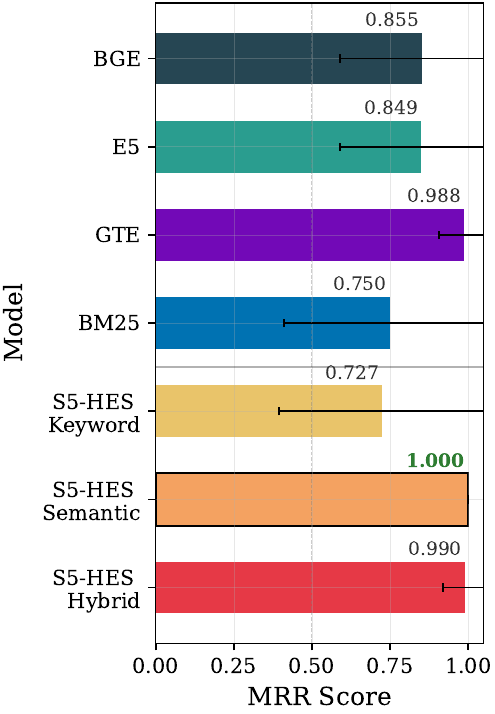} 
\label{eval_fig_2_A}
\end{minipage}%
}%
\subfloat[Mean average precision vs Models]{
\begin{minipage}[t]{0.50\linewidth} 
\hspace{-7mm}  
\centering 
\includegraphics[scale=0.5]{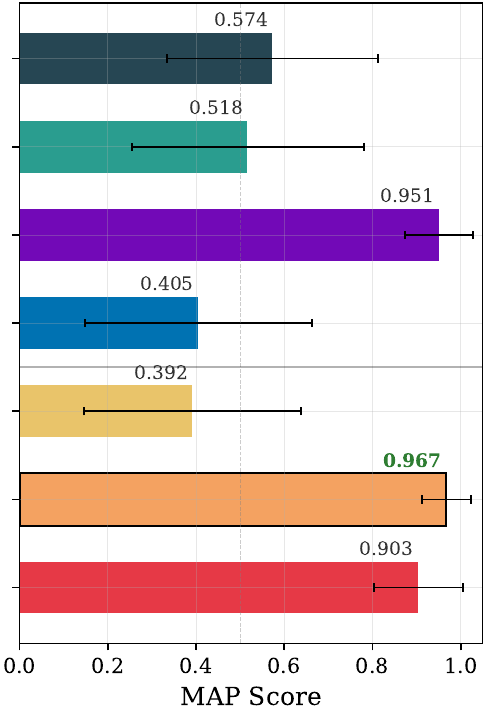}
\label{eval_fig_2_B}
\end{minipage} 
}
\caption{Position-based metrics comparison 
\newline (Error bars show the std of per-query scores across the 100 queries.)
}
\vspace{-5mm}
\label{eval_fig_2}
\end{figure}

Results are presented in three stages following the standard information retrieval evaluation hierarchy. First, set-based metrics (Precision@K, Recall@K) and rank-weighted quality (nDCG@K) characterize how retrieval performance evolves with the number of returned documents, as shown in Fig. \ref{eval_fig_1}. 
Second, position-based summary metrics (MRR, MAP) condense retrieval effectiveness into single scores that are directly comparable across models, as shown in Fig. \ref{eval_fig_2}. Table \ref{eval_table_1} presents the latency analysis
of each retrieval strategy, assessing quality gains and overhead.

Fig. \ref{eval_fig_1} presents three distinct performances against all metrics. 
S5-HES Semantic and GTE-based form the top tier, with S5-HES Semantic achieving perfect P@1 = 1.000 and both models reaching R@20 = 1.000, indicating that all relevant documents are retrieved within the top 20 results. 
S5-HES Hybrid, BGE-base, and E5-base occupy the middle tier, while BM25 and S5-HES Keyword cluster at the bottom with near-identical performance (e.g., P@10 = 0.291 for both). 
The classic precision-recall trade-off is clearly visible between Fig. \ref{eval_fig_1_A} and Fig. \ref{eval_fig_1_B}: as K increases from 1 to 20, precision declines monotonically (S5-HES Semantic: 1.000 to 0.301) while recall rises (0.203 to 1.000).
Fig. \ref{eval_fig_1_C} shows that nDCG@K remains notably stable for the top-tier models, S5-HES Semantic maintains 0.975-1.000 across all K values, confirming that relevant documents are consistently ranked near the top regardless of cutoff depth. 
S5-HES Hybrid ranks third despite fusing keyword and semantic retrieval via reciprocal rank fusion, and its keyword component introduces noise that dilutes precision relative to the pure semantic approach. 

Fig. \ref{eval_fig_2} shows retrieval quality into single per-model scores, revealing complementary insights. 
Fig. \ref{eval_fig_2_A} shows that MRR compresses differences at the top: S5-HES Semantic achieves a perfect MRR of 1.000, meaning the first relevant document is ranked first for every query, while S5-HES Hybrid (0.990) and GTE-base (0.988) are nearly indistinguishable. 
Fig. \ref{eval_fig_2_B} presents differences between MAP, which evaluates precision at every relevant document position rather than only the first. 
Here, a clear separation emerges. 
S5-HES Semantic (0.967) and GTE-base (0.951) lead.
But S5-HES Hybrid drops to 0.903, a gap of 0.087 from its MRR score, indicating that its top-ranked result is almost always relevant.
%
Below these three models, a cliff separates the top tier from the remaining baselines: BGE (0.574) and E5 (0.518) achieve moderate MAP scores, while BM25 (0.405) and S5-HES Keyword (0.392) occupy the bottom tier with the widest error bars, reflecting inconsistent lexical matching performance across the five smart home knowledge categories. 
Taken together, MRR and MAP confirm that S5-HES Semantic delivers strongest retrieval quality by both first-hit and comprehensive ranking criteria.

\begin{table}[t]
\vspace{-5mm}
  \centering
  \caption{Query latency summary (ms) over 100 queries}
  \label{tab:latency}
  \begin{tabular}{lrrr}
    \hline
    \textbf{Model} & \textbf{Median} & \textbf{Mean} & \textbf{P95} \\
    \hline
    S5-HES Hybrid   & 204.4 & 313.3 & 225.1 \\
    S5-HES Semantic & 136.3 & 136.8 & 153.5 \\
    S5-HES Keyword  &  34.2 &  34.9 &  48.8 \\
    \hline
    BM25            &  42.9 &  43.9 &  54.3 \\
    GTE             & 107.5 & 107.4 & 122.0 \\
    E5              & 120.4 & 137.1 & 141.7 \\
    BGE             & 177.7 & 199.2 & 196.2 \\
    \hline
  \end{tabular}
\vspace{-7mm}
\label{eval_table_1}
\end{table}

Table \ref{eval_table_1} shows that all retrieval methods, across all models, achieve P95 $<$ 230 ms. 
S5-HES Keyword (median 34.2 ms) and BM25 (42.9 ms) are the fastest, as expected for lexical matching. 
Semantic models occupy a middle band and S5-HES Semantic (136.3 ms), GTE (107.5 ms), and E5 (120.4 ms). 
S5-HES Hybrid has the highest median latency (204.4 ms) due to executing both keyword and semantic retrieval, followed by reciprocal rank fusion, with an elevated mean (313.3 ms) relative to its P95 (225.1 ms), indicating occasional outlier queries with higher processing time.
Meanwhile, S5-HES Semantic delivers the highest retrieval quality (MRR=1.000 and MAP=0.967) at a median latency of 136.3 ms, offering the best quality-to-latency trade-off among all evaluated models.

\subsubsection{Response quality}
\begin{figure}[t]
\vspace{-5mm}
\centering
\includegraphics[width=\columnwidth]{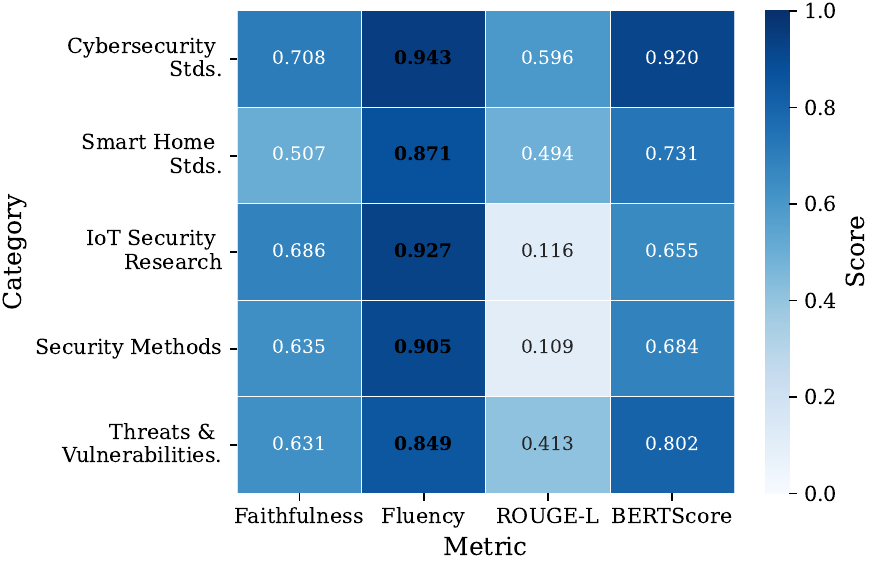}
\caption{Category-wise generation quality}
\label{eval_fig_3}
\vspace{-9mm}
\end{figure}

\begin{figure}[t]
\vspace{-5mm}
\centering
\includegraphics[width=\columnwidth]{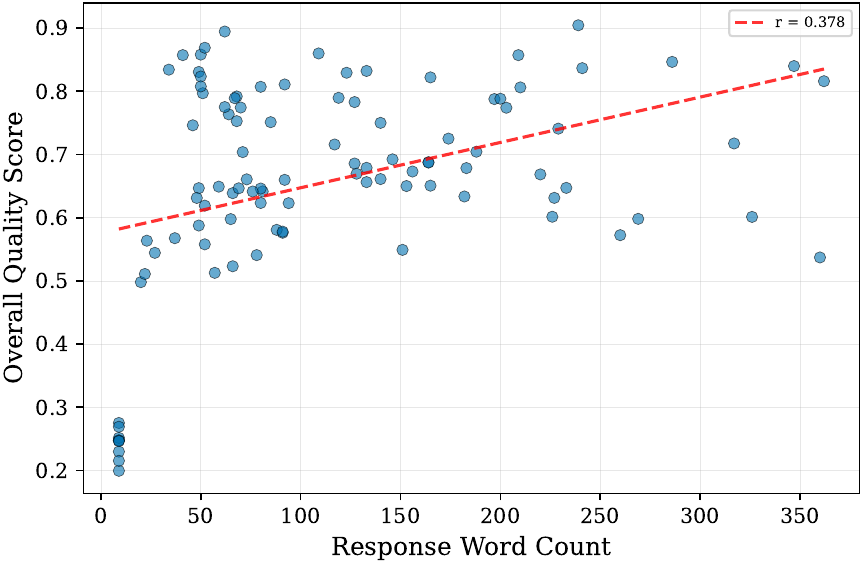}
\caption{Response length vs quality correlation}
\label{eval_fig_4}
\vspace{-8mm}
\end{figure}

Response quality is evaluated mainly using four core metrics: faithfulness, fluency, ROUGE-L, and BERTScore. Employ 100 queries under five smart home knowledge categories across three LLMs that share the same retrieval pipeline.
Fig. \ref{eval_fig_3} present per-category heatmap that reveals domain-dependent quality variation. Fig. \ref{eval_fig_4} shows a scatter plot examining whether response length confounds quality scores, and Table \ref{eval_table_2} presents a provider comparison that isolates the effect of the generation model from the retrieval stage.

Fig. \ref{eval_fig_3} reveals notable category-dependent variation. Fluency is uniformly high across all categories, confirming structurally coherent generation across domains. The primary differentiator is ROUGE-L, which ranges from 0.596 (Cybersecurity Standards) to 0.109 (Security Methods), a time gap reflecting the lexical diversity of open-ended research literature versus well-structured standards documents. 
Cybersecurity Standards achieves the strongest overall profile, while IoT Security Research and Security Methods exhibit low ROUGE-L scores despite maintaining moderate Faithfulness, indicating that the model captures semantic intent but paraphrases 
occasionally.




Fig. \ref{eval_fig_4} shows a moderate positive correlation between response word count and overall quality. The relationship is partly driven by a cluster of very short responses ($<$25 words) scoring below 0.30, which likely represent incomplete generations. 
Beyond this floor effect, responses in the 50-200 word range exhibit wide vertical spread (0.45-0.88), demonstrating that length alone does not determine quality; concise, well-grounded responses can match or exceed longer ones. Even the longest responses ($>$300 words) scatter between 0.50 and 0.85, confirming that verbosity provides no quality guarantee. This suggests retrieval precision, not response length, is the primary driver of generation quality.

According to Table \ref{eval_table_2}, Gemini 2.0 Flash achieves the highest overall quality and the lowest cloud latency.
GPT-4o attains the highest Fluency but lower ROUGE-L, suggesting more paraphrased outputs.
Llama 3.2 3B trails overall quality metrics.
Notably, Fluency and Faithfulness remain comparable across all three providers; the main differentiator is ROUGE-L, indicating that the shared retrieval pipeline, rather than the generation model, is the dominant factor.

\begin{table}[t]
\centering
\caption{Multi-LLM provider comparison}
\label{tab:llm_comparison}
\begin{tabular}{lccccc}
\hline
\textbf{Provider} & \textbf{Faith} & \textbf{Fluency} & \textbf{ROUGE} & \textbf{BERT} & \textbf{Latency} \\
\hline
Gemini 2.0 Flash & 0.774 & 0.948 & 0.556 & 0.900 & 3,129 \\
GPT-4o           & 0.720 & 0.986 & 0.347 & 0.830 & 4,194 \\
Llama 3.2 3B     & 0.683 & 0.969 & 0.265 & 0.758 & 154,267 \\
\hline
\label{eval_table_2}
\vspace{-9mm}
\end{tabular}
\end{table}

\subsection{Data generation engine}
\label{eval_sec_4}
This subsection organized around four questions.
First, do the simulated threat scenarios look like real attacks? 
Results are presented in Section \ref{eval_sub2_sub1}.
The second question is, do the generated messages resemble real IoT telemetry? 
and Section \ref{eval_sub2_sub2} discussed the results.
As the third, how does the overall dataset hold up against established benchmarks? 
In Section \ref{eval_sub2_sub3}, we discussed the experiment and the results.
Finally, the fourth is, what can S5-HES do that static datasets cannot? 
and Section \ref{eval_sub2_sub4} elaborated finding on that.

\subsubsection{Threat scenario fidelity}
\label{eval_sub2_sub1}

\begin{figure}[t]
\vspace{-9mm}
\hspace{-9mm}  
\subfloat[Attack behaviour coverage]{
\begin{minipage}[t]{0.51\linewidth} 
\includegraphics[scale=0.5]{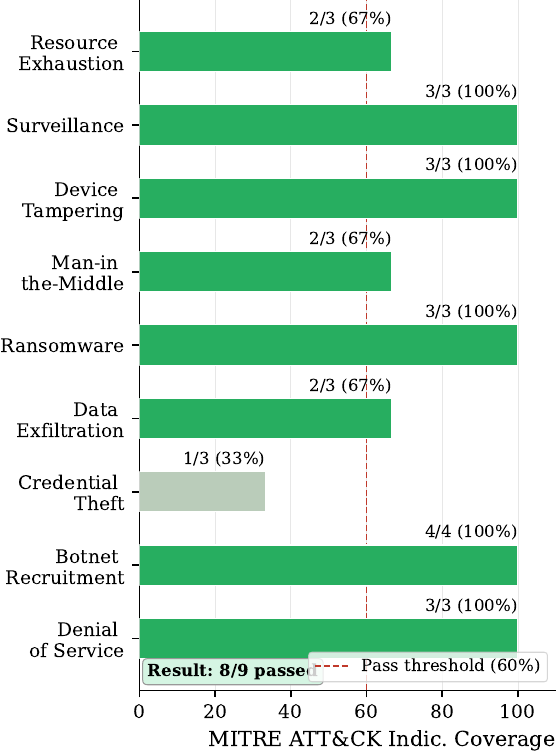} 
\label{eval_fig_5_A}
\end{minipage}%
}%
\subfloat[Attack lifecycle fidelity]{
\begin{minipage}[t]{0.48\linewidth} 
\includegraphics[scale=0.5]{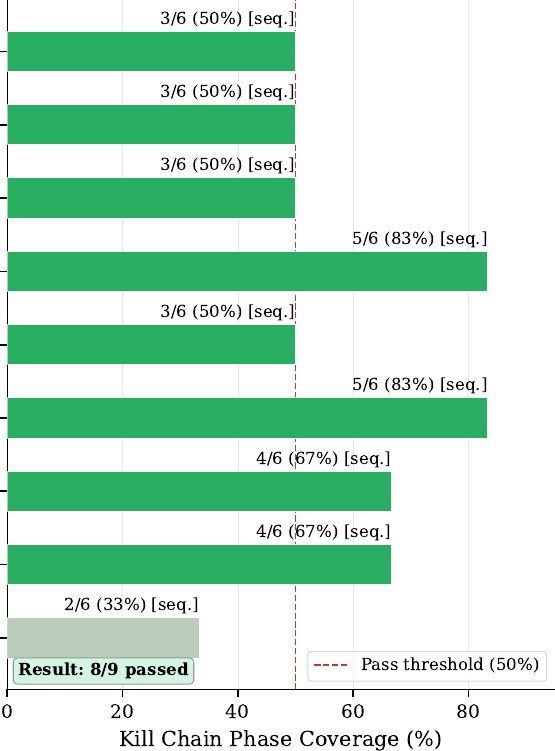}
\label{eval_fig_5_B}
\end{minipage} 
}
\caption{Position-based metrics comparison 
}
\label{eval_fig_5}
\vspace{-9mm}
\end{figure}

We evaluate threat scenario fidelity by comparing S5-HES simulated attack lifecycles with labelled network traffic from three baseline datasets, Edge-IIoTset, IoT-23, and Bot-IoT, across nine overlapping threat types. 
Each simulated scenario follows a multi-phase lifecycle (e.g., reconnaissance, exploitation, lateral movement, exfiltration) governed by configurable timing and event probability parameters; all runs use a fixed seed for reproducibility, and baseline samples are capped at 5,000 per threat.
We evaluate threat scenario fidelity using two metrics, attack behaviour coverage (ABC) and attack lifecycle fidelity (ALF), and present the results in Fig. \ref{eval_fig_5}.
%
%
Fig. \ref{eval_fig_5_A} presents ABC measures MITRE ATT$\&$CK indicator coverage (pass: $\geq$60$\%$), and ALF measures cyber kill chain phase coverage with valid sequential ordering (pass: $\geq$50$\%$) and presented in Fig. \ref{eval_fig_5_B}.

According to the Fig. \ref{eval_fig_5_A}, eight of nine threats pass the 60$\%$ threshold. Four threats, denial of service, botnet recruitment, man-in-the-middle, and surveillance, achieve full indicator coverage (3/3, 100$\%$). Resource exhaustion, data exfiltration, device tampering, and ransomware each match two of three indicators (2/3, 67$\%$). Credential Theft is the sole failure (1/3, 33$\%$), matching only one of three expected MITRE ATT$\&$CK indicators.

As shown in Fig. \ref{eval_fig_5_B}, eight of nine threats pass the 50$\%$ phase coverage threshold. All nine exhibit valid sequential ordering. Man-in-the-middle and data exfiltration achieve the highest coverage (5/6, 83$\%$), followed by botnet recruitment and credential theft (4/6, 67$\%$). Four threats, ransomware, surveillance, device tampering, and resource exhaustion sit at the boundary (3/6, 50$\%$). Denial of service is the sole failure (2/6, 33$\%$), covering only two of six kill chain phases. Notably, the two metrics identify different weaknesses: credential theft fails ABC but passes ALF, while denial-of-service fails ALF but passes ABC.

Together, ABC and ALF confirm that the simulated threats are both behaviourally accurate (reproducing expected MITRE ATT$\&$CK indicators) and structurally sound (traversing kill chain phases in valid order). Eight of nine threats pass both metrics. Each failures reflect inherent characteristics of those attack types rather than general simulation deficiencies. 
Denial of service is a volumetric attack that floods targets without progressing through a full kill chain, so its lifecycle naturally covers fewer phases. Credential theft in the baseline (Edge-IIoTset) is characterised by network-level brute-force patterns, whereas S5-HES simulates credential theft at application layer. 

\subsubsection{Device behavior realism}
\label{eval_sub2_sub2}
To validate the realism of S5-HES simulated device behaviour, we employ the message similarity (M.SIM) metric, which quantifies structural and semantic alignment between generated IoT messages and real-world telemetry from two baselines, SDHAR-HOME and Logging. M.SIM comprises four sub-metrics, field coverage, type compatibility, range overlap, and semantic similarity, and is aggregated as an equal-weight mean. Evaluation is conducted across 23 overlapping devices, using three comparison modes: vs SDHAR, vs Logging, and Hybrid (combined) and results are shown in Table \ref{eval_table_3}.

Field coverage is perfect across all comparisons (100$\%$), confirming that S5-HES messages contain every expected field. Type compatibility exceeds 90$\%$ in all modes, with minor mismatches in approximately 9$\%$ of fields. Semantic similarity is consistently high (94.3  95.4$\%$), indicating that generated values carry correct meaning.
Range overlap is the weakest sub-metric (66.7-75.6$\%$), as simulated sensor readings do not fully replicate real-world value distributions.
Logging scores lowest (66.7$\%$) due to its narrower recording window ($\sim$ 15 days) compared to SDHAR-HOME (62 days, 33 sensors). Combined M.SIM scores of 88.4  90.0$\%$ across all comparison modes confirm that S5-HES produces structurally complete and semantically accurate IoT messages.

\subsubsection{Data quality}
\label{eval_sub2_sub3}
\begin{figure}[H]
\vspace{-3mm}
\centering
\includegraphics[width=\columnwidth]{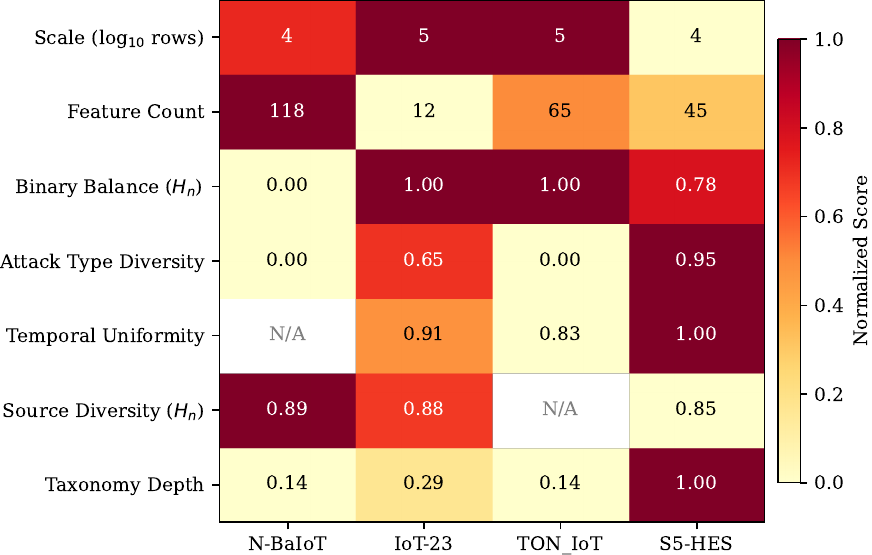}
\caption{Category-wise generation quality}
\label{eval_fig_7}
\vspace{-5mm}
\end{figure}

S5-HES benchmark against three established smart home and IoT security datasets, N-BaIoT, IoT-23, and TON-IoT across seven quality metrics;
dataset scale (D1), feature count (D2), binary class balance (D3), attack type diversity (D4), temporal uniformity (D5), source/device diversity (D6), and label taxonomy depth (D7). 
Each metric is computed identically across all four datasets and min-max normalised per row, so that the lowest-scoring dataset maps to 0 and the highest to 1. 
Not all metrics apply to every dataset: N-BaIoT lacks timestamps (D5 unavailable), and TON-IoT lacks a usable device column (D6 unavailable). 
Fig. \ref{eval_fig_7} heatmap summarises the comparison, with raw values annotated in cells.

\begin{table}[t]
\vspace{-9mm}
  \centering
  \caption{Message similarity against SDHAR and Logging}
  \label{eval_table_3}
  \begin{tabular}{lccc}
    \hline
    \textbf{Metric} & \textbf{SDHAR} & \textbf{Logging} & \textbf{Hybrid} \\
    \hline
    Field Coverage    & 100.0\% & 100.0\% & 100.0\% \\
    Type Compatibility      &  90.0\% &  91.7\% &  91.7\% \\
    Range Overlap     &  75.6\% &  66.7\% &  73.2\% \\
    Semantic Similarity          &  94.3\% &  95.4\% &  95.3\% \\
    \hline
    \textbf{Combined}   &  90.0\% &  88.4\% &  90.0\% \\
    \hline
  \end{tabular}
  \vspace{-6mm}
\end{table}

S5-HES achieves the highest normalised score on three of seven metrics. Attack type diversity (0.95) exceeds all baselines, N-BaIoT and TON-IoT score 0.00 (single attack type or no differentiation), and IoT-23 reaches 0.65. 
Temporal uniformity is 0.99, above IoT-23 (0.91) and TON-IoT (0.83). 
Taxonomy depth reaches the maximum (1.00), reflecting a seven-level hierarchical label structure; the baselines range from 0.14 to 0.29 (one to two levels). 
Binary balance is reasonable (0.78), though below the perfect balance of IoT-23 and TON-IoT (both 1.00), N-BaIoT is significantly imbalanced (0.00).
Source diversity is comparable across datasets (S5-HES 0.85, N-BaIoT 0.89, IoT-23 0.88).
S5-HES scores lowest on the dataset scale 
compared to IoT-23 and TON-IoT 
and its feature count (45) is mid-range between IoT-23 (12) and N-BaIoT (118). The scale gap is expected; this experiment uses a single simulation run, and S5-HES can generate arbitrarily large datasets by increasing the number of homes or duration, as demonstrated in Section \ref{eval_sub2_sub4}.


\subsubsection{Dataset Capabilities}
\label{eval_sub2_sub4}
\begin{figure}[!t]
\vspace{-9mm}
\hspace{-10mm}  
\centering
\subfloat[Total devices]{
\begin{minipage}[t]{0.33\linewidth} 
\centering 
\includegraphics[scale=0.35]{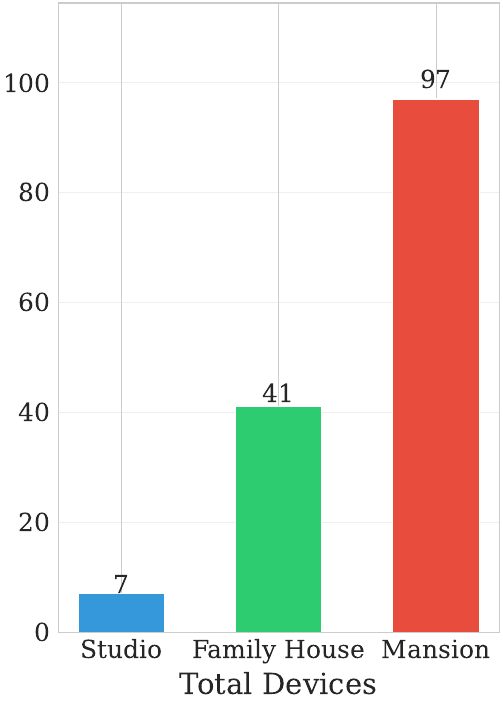} 
\label{eval_fig_8_A}
\end{minipage}%
}%
\subfloat[Filtered events (24h)]{
\begin{minipage}[t]{0.31\linewidth} 
\centering 
\includegraphics[scale=0.35]{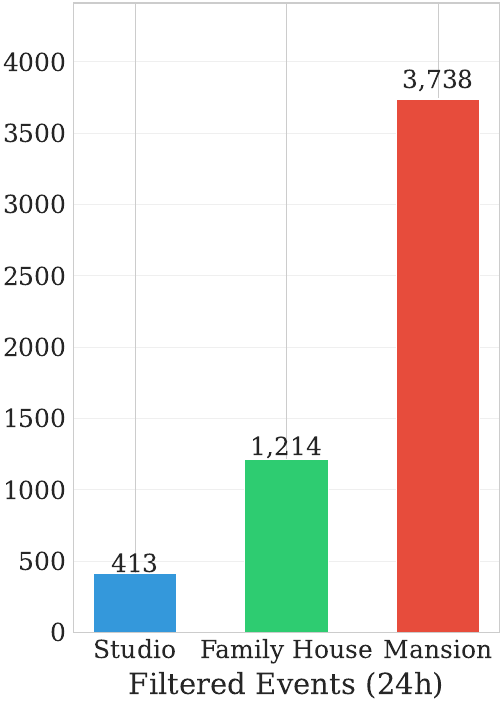}
\label{eval_fig_8_B}
\end{minipage} 
}
\subfloat[Unique device ]{
\begin{minipage}[t]{0.27\linewidth} 
\centering 
\includegraphics[scale=0.33]{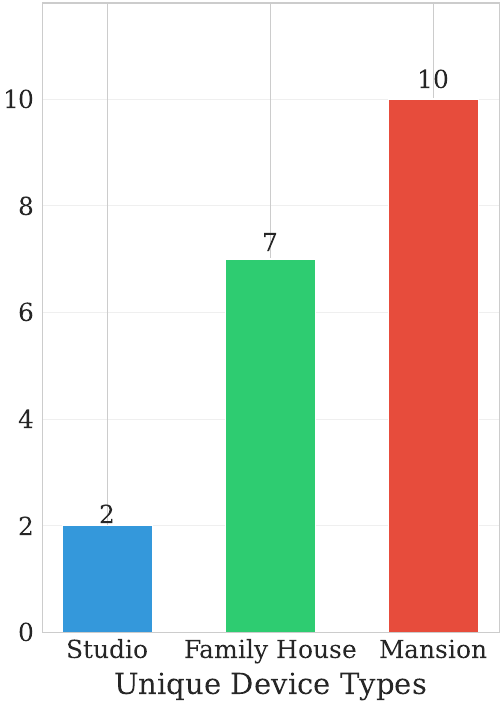}
\label{eval_fig_8_C}
\end{minipage} 
}
\caption{S5-HES scalability across residence configs
}
\label{eval_fig_8}
\vspace{-9mm}
\end{figure}

\begin{figure}[t]
\vspace{-9mm}
\centering
\subfloat[Event distribution diversity]{
\begin{minipage}[t]{0.50\linewidth} 
\centering 
\includegraphics[scale=0.32]{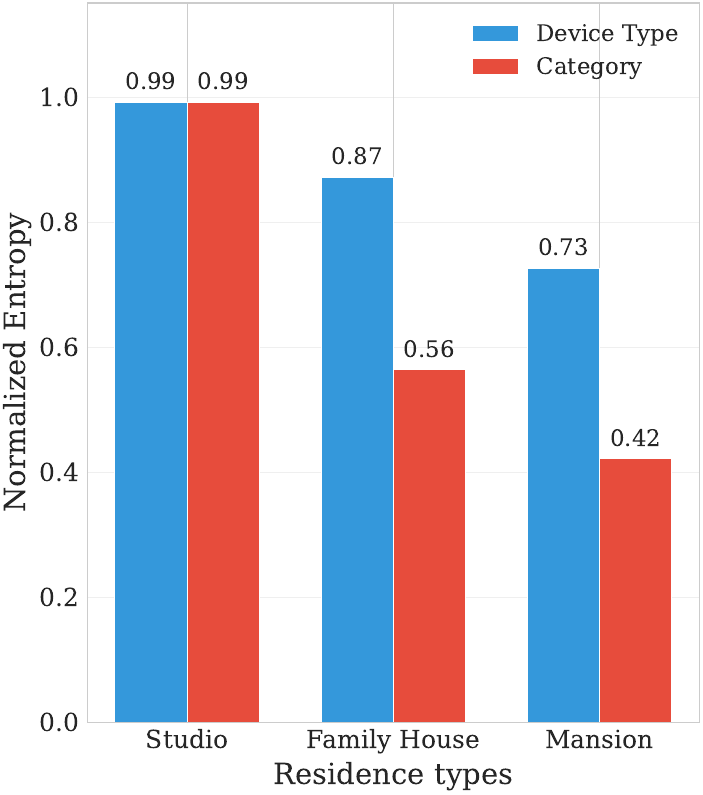} 
\label{eval_fig_10_A}
\end{minipage}%
}%
\subfloat[Event distribution inequality]{
\begin{minipage}[t]{0.50\linewidth} 
\centering 
\includegraphics[scale=0.32]{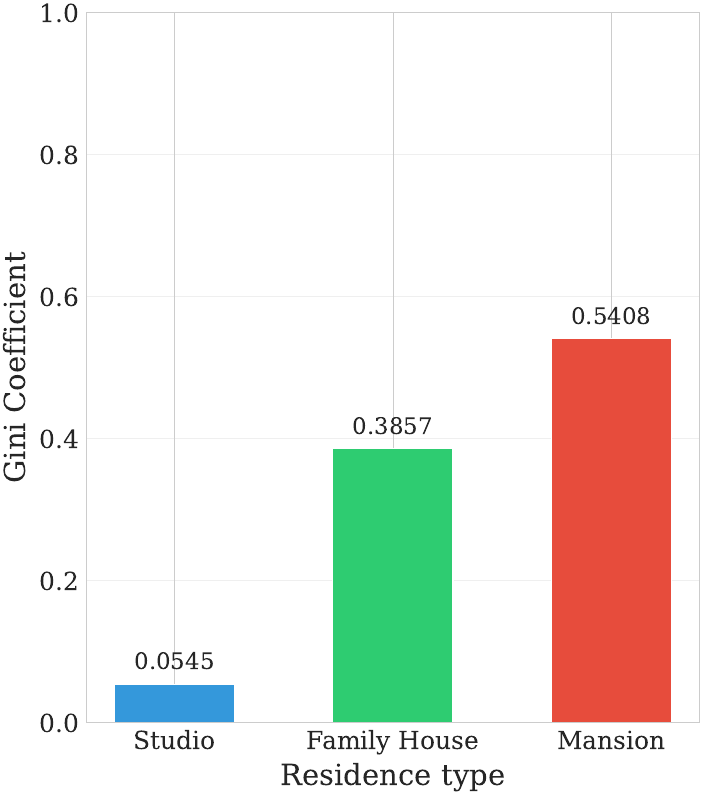}
\label{eval_fig_10_B}
\end{minipage} 
}
\caption{Device diversity across home template scales}
\vspace{-9mm}
\label{eval_fig_10}
\end{figure}

We demonstrate two key capabilities that static benchmark datasets cannot offer: configurable scalability and preservation of device diversity across scales. Three home templates of increasing complexity: Studio (3 rooms, 7 devices, 1 inhabitant), Family House (13 rooms, 41 devices, 4 inhabitants), and Mansion (21 rooms, 97 devices, 6 inhabitants) are each simulated for 24 hours under identical parameters with normal traffic only. Fig. \ref{eval_fig_8} shows how total devices, filtered event volume, and unique device types scale with template complexity. 
Fig. \ref{eval_fig_10} measures device diversity at each scale using normalised Shannon entropy (higher = more diverse) and Gini coefficient (lower = more balanced).

All three metrics increase monotonically with template complexity. Total devices grow from 7 (Studio) to 41 (Family House) to 97 (Mansion), a 13.9$\times$ increase. Filtered event volume scales from 413 to 1,223 to 3,742 (9.1$\times$), and unique device types from 2 to 7 to 10 (5$\times$). Events per device vary across templates (59.0, 29.8, 38.6) rather than remaining constant, reflecting the changing device mix: larger homes introduce more passive devices (e.g., water leak sensors, smoke detectors) that generate fewer events than active devices (eg. motion sensors, routers). Since the only parameter changed is the home template, these results confirm that S5-HES scales predictably by configuration alone.

Device type entropy decreases from 0.99 (Studio) to 0.88 (Family House) to 0.74 (Mansion), and category entropy follows the same trend (0.99, 0.57, 0.43). The Gini coefficient rises correspondingly from 0.0683 to 0.3728 to 0.5306. This pattern is expected: the Studio has only two device types sharing events nearly equally, yielding near-perfect entropy and minimal inequality. Larger templates introduce more device types with inherently different activity rates. Routers and motion sensors generate events far more frequently than smoke detectors or water leak sensors, producing a natural imbalance. This mirrors real smart home environments where event distribution across device types is uneven by design, not a simulation artifact. The results demonstrate that S5-HES maintains realistic device heterogeneity as the generated environment scales up.
\subsection{Discussion}
\label{eval_sec_5}
The discussion is conducted under three subsections: key findings, limitations, and threats to validity.
\subsubsection{Key Findings}
The S5-HES retrieval pipeline achieves higher document retrieval accuracy. S5-HES Semantic places the first relevant document at rank one for queries, and maintains MAP=0.967 across all relevant positions. GTE-base is the closest baseline. The Hybrid variant, which fuses keyword and semantic retrieval, ranks third. The keyword component introduces ranking noise, reducing average precision. 
Fluency and Faithfulness remain comparable across LLM providers, suggesting the shared retrieval pipeline is a stronger determinant of quality than the generation model.

Threat scenario evaluation shows that S5-HES reproduces expected attack behaviours for the majority of tested threats. 
ABC and ALF each yield eight of nine threats that pass both metrics. The two failures involve threats: Credential Theft matches only 1/3 MITRE ATT$\&$CK indicators despite following a valid lifecycle (4/6 phases), and Denial of Service covers 2/6 Kill Chain phases, consistent with its volumetric nature. All nine threats exhibit valid sequential ordering in ALF, indicating that the simulation engine preserves attack lifecycle structure even where phase coverage is incomplete.

Device behaviour comparison against SDHAR-HOME and Logging yields combined M.SIM scores of 88.4-90.0$\%$. Field coverage is appreciably high
and semantic similarity is high.
S5-HES scores highest on attack type diversity, temporal uniformity, and taxonomy depth, while scoring lowest on dataset scale
The scale gap reflects the single-run configuration used in this evaluation; the scalability results shown in Section \ref{eval_sub2_sub4} suggest that larger datasets can be achieved by adjusting template parameters, though generation at baseline-matching volumes has not been demonstrated.

Scalability experiments confirm that the S5-HES dataset size grows predictably with template complexity: devices scale, filtered events, and unique device types from Studio to Mansion. Device diversity metrics show that event distribution inequality increases at larger scales (Gini: 0.07 to 0.53), expected in real smart home environments where different device types have inherently different activity rates.

\subsubsection{Limitations}

Several limitations should be noted. First, the M.SIM evaluation covers only 
overlapped limited device types because of the limitations of SDHAR-HOME and Logging.
Second, all data generation experiments use a single random seed. While this ensures reproducibility, it does not demonstrate robustness across different random initialisations; a multi-seed evaluation would strengthen confidence in the results. Third, the ROUGE-L score for generation quality falls. This reflects paraphrasing behaviour rather than factual inaccuracy. Thus, this indicates a gap between the expected answer format and the LLM's output style. 
Fifth, Credential Theft matches only one of three expected MITRE ATT$\&$CK indicators, suggesting the simulation does not yet capture all catalogued attack behaviours for this threat type. Finally, the quality heatmap comparison presented in Section \ref{eval_sub2_sub3} uses min-max normalisation across four datasets, meaning scores are relative to the specific dataset pool and would change if additional datasets were included.

\subsubsection{Threats to Validity}
discussed in three states as follows.

\textit{Internal validity}: The evaluation metrics (M.SIM, ABC, ALF) are implemented and unit-tested within the project.
The generation quality metrics (Faithfulness, Fluency) use embedding-based scoring, which may not capture all dimensions of response quality.

\textit{External validity}: The baseline datasets represent specific IoT environments and capture conditions. SDHAR-HOME and Logging reflect two largely distinct resident configurations; the Edge-IIoT set, IoT-23, and Bot-IoT record traffic from specific network testbeds. Results may not generalise to all smart home deployments or network environments. Only three home templates were tested for scalability; intermediate or more extreme configurations remain unevaluated.

\textit{Construct validity}: ABC relies on keyword matching against MITRE indicator lists may miss semantically equivalent attack behaviors expressed in different terminology. ALF assumes a six-phase Cyber Kill Chain model; attacks that do not follow linear structure may be at a disadvantage. ROUGE penalizes valid paraphrasing, potentially underestimate generation quality for responses, which semantically correct but lexically differ from the referred answers.

\section{Conclusion}
\label{sec_7}
To the best of our knowledge, S5-HES Agent is the first automated, AI-augmented framework with agentic orchestration designed to democratize smart home environment simulation. 
The framework integrates a RAG-enhanced knowledge base, multi-agent orchestration with specialized agents for home configuration, device management, and threat injection, and a human-in-the-loop verification pipeline ensuring research integrity. 

Despite these strengths, limitations remain in the value-range overlap with real-world sensor data, lexical alignment in generated responses, and the scope of device behaviour validation, constrained by available baseline datasets. 
Future work will address these gaps through multi-seed robustness evaluation, expanded real-world device validation, and the integration of adaptive agent reasoning to further advance autonomous operation. 
We plan to extend S5-HES Agent toward a digital twin, enabling real-time synchronization between simulated and physical smart home environments, and to explore metaverse integration for immersive, interactive simulation and training scenarios. 
The S5-HES Agent framework is publicly available under the MIT License at https://github.com/AsiriweLab/S5-HES-Agent.




\bibliographystyle{IEEEtran}
\bibliography{reference}


\begin{IEEEbiography}[{\includegraphics[width=1in,height=1.25in,clip,keepaspectratio]{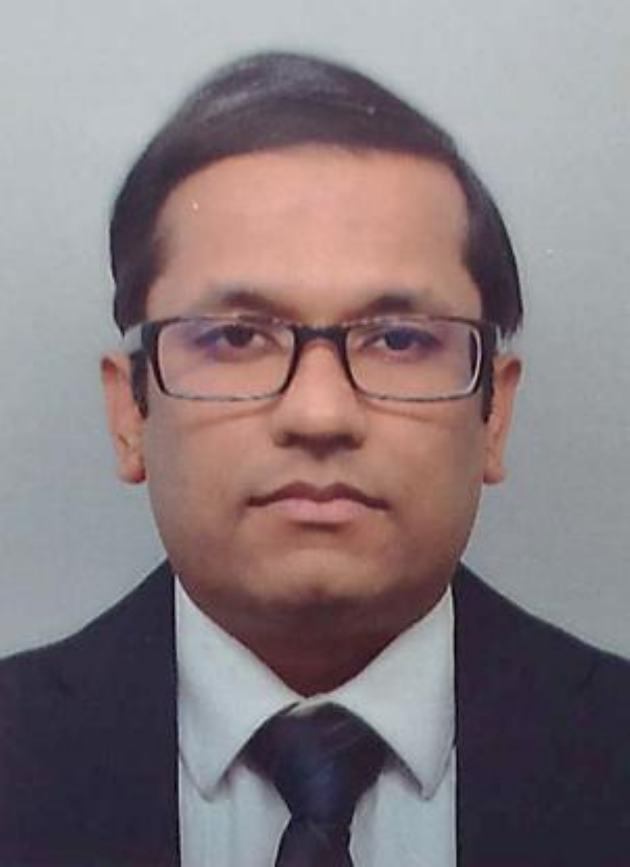}}]{Akila Siriweera}
is an associate professor at the University of Aizu. He received BSc from the University of Peradeniya, Sri Lanka and an MSc and PhD in computer science and engineering from the University of Aizu, Japan. His current research interests include Agentic AI, Big Data, and Web 3.0. He has received several outstanding awards in academia and the industry. He is an IEEE member and a TC member of the IEEE Consumer Technology Society IoT group.
\end{IEEEbiography}

\vspace{-5mm}

\begin{IEEEbiography}[{\includegraphics[width=1in,height=1.25in,clip,keepaspectratio]{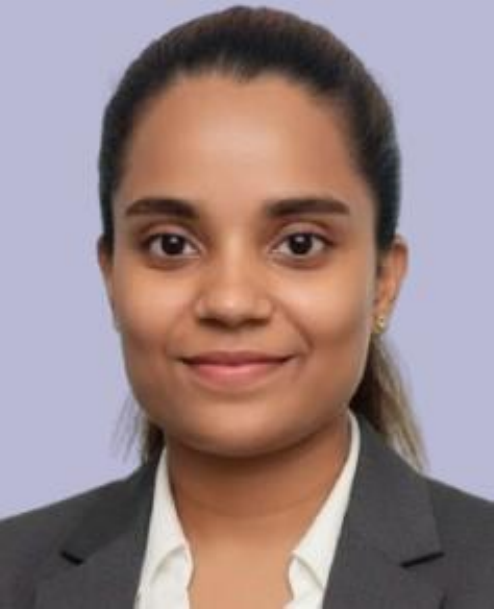}}]{Janani Rangila} is in her final year of her BSc at the Information Technology Faculty at KD University, Sri Lanka.
She is working in AI, distributed computing, and mobile applications. 
Her research interests include Agentic AI and Web 3.0. 
She has been working on her final year research, which is an extension of this work.
\end{IEEEbiography}

\vspace{-5mm}

\begin{IEEEbiography}[{\includegraphics[width=1in,height=1.25in,clip,keepaspectratio]{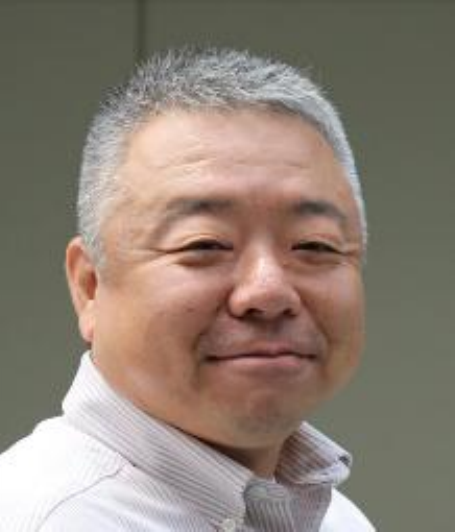}}]{Keitaro Naruse}
is a professor at the University of Aizu, Japan. He has specialized in swarm robots and applications for agricultural  robotic systems and robot interface systems in disaster responses. 
He works for design, DevOps, and standardized networked  distributed intelligent robot systems with heterogeneous sensors and robots. 
His research team has received several awards in various international robot competitions.
\end{IEEEbiography}

\vspace{-5mm}

\begin{IEEEbiography}[{\includegraphics[width=1in,height=1.25in,clip,keepaspectratio]{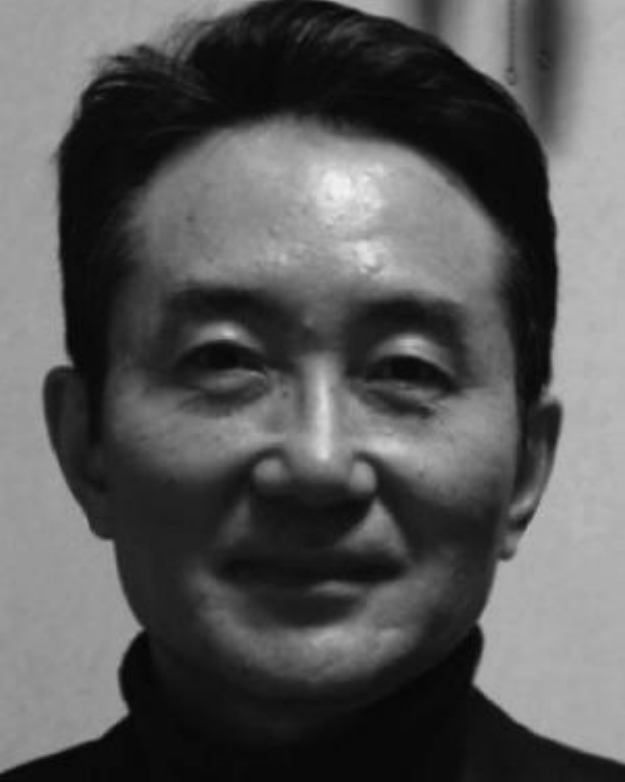}}]{Incheon Paik}
(Senior Member, IEEE) received the M.E. and Ph.D. degrees in electronic engineering from Korea University in 1987 and 1992, respectively. He is currently a professor with the University of Aizu, Japan. His research interests include deep learning applications, ethical LLMs, machine learning, big data science, and semantic web services. He is a member of the IEICE, IEIE, and IPSJ.
\end{IEEEbiography}

\vspace{-5mm}

\begin{IEEEbiography}[{\includegraphics[width=1in,height=1.25in,clip,keepaspectratio]{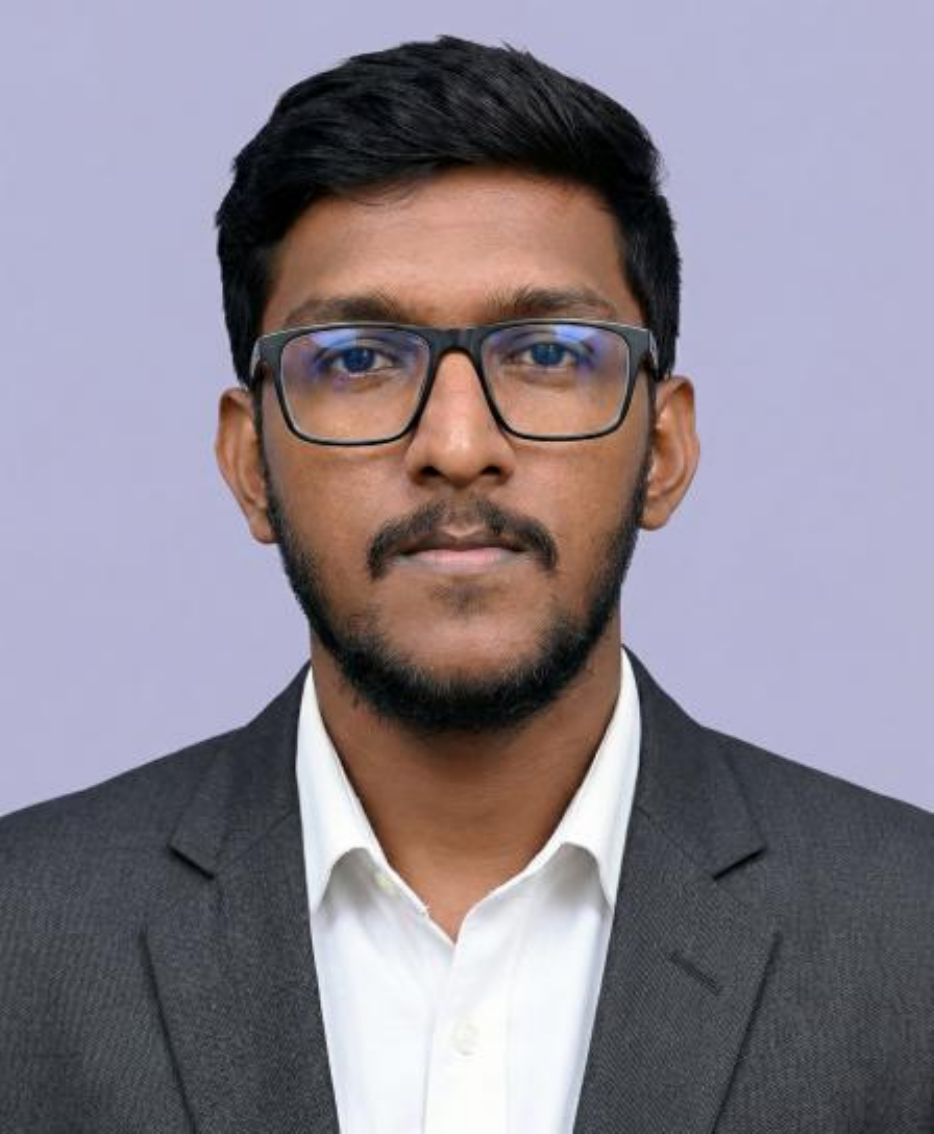}}]{Isuru Jayanada} is an undergraduate student at KD University with a strong interest in Agentic AI, full-stack software development, and information technology. He has practical experience in designing and developing web and mobile applications, with a focus on creating efficient, user-centered solutions. His academic and project work reflect a commitment to advancing his technical skills and contributing to the field of computing.
\end{IEEEbiography}

\vfill\pagebreak

\end{document}